\documentclass[10pt,twocolumn,letterpaper]{article}

\usepackage{mysty}
\usepackage{times}
\usepackage{epsfig}
\usepackage{graphicx}
\usepackage{amsmath}
\usepackage{amssymb}

\usepackage{multirow}
\usepackage{booktabs}
\usepackage{siunitx}
\usepackage{subcaption}
\usepackage{bigstrut}
\usepackage{float}
\usepackage{setspace}

\usepackage{appendix}

\usepackage{tikz}
\usetikzlibrary{decorations.pathmorphing}
\usetikzlibrary{intersections}
\usetikzlibrary{positioning}
\usetikzlibrary{calc}
\usetikzlibrary{fit}
\usepackage[skins]{tcolorbox}

\usepackage{upgreek}
\usepackage{bm}
\makeatletter
\newcommand{\bfgreek}[1]{\bm{\@nameuse{#1}}}
\newcommand{\bfgreekcolor}[2]{\bm{\textcolor{#1}{\@nameuse{#2}}}}
\makeatother

\usepackage{cvpr}

\newif\ifshowcomment
\showcommenttrue
\newcommand{\xxcom}[2]{\ifshowcomment\textcolor{#1}{#2}\fi}
\newcommand{\rtcom}[1]{\xxcom{blue}{#1}} 
\newcommand{\bpcom}[1]{\xxcom{green}{#1}} 

\newcommand{\supporapp}[0]{\ifarxiv Appendix\else supplementary material\fi}

\newif\ifarxiv
\arxivtrue

\usepackage[pagebackref=true,breaklinks=true,letterpaper=true,colorlinks,bookmarks=false]{hyperref}

\cvprfinalcopy 

\def\cvprPaperID{1946} 

\ifcvprfinal\fi
\begin{document}

\title{Discriminability objective for training descriptive captions} 

\author{Ruotian Luo\\
TTI-Chicago\\
{\tt\small rluo@ttic.edu}
\and
Brian Price\\
Adobe Research\\
{\tt\small bprice@adobe.com}
\and
Scott Cohen\\
Adobe Research\\
{\tt\small scohen@adobe.com}
\and
Gregory Shakhnarovich\\
TTI-Chicago\\
{\tt\small greg@ttic.edu}
}

\maketitle


\begin{abstract}
One property that remains lacking in image captions generated by contemporary methods is discriminability: being able to tell two images apart given the caption for one of them. We propose a way to improve this aspect of caption generation. By incorporating into the captioning training objective a loss component directly related to ability (by a machine) to disambiguate image/caption matches, we obtain systems that produce much more discriminative caption, according to human evaluation. Remarkably, our approach leads to improvement in other aspects of generated captions, reflected by a battery of standard scores such as BLEU, SPICE etc. Our approach is modular and can be applied to a variety of model/loss combinations commonly proposed for image captioning. Code has been made available at: \href{https://github.com/ruotianluo/DiscCaptioning}{https://github.com/ruotianluo/DiscCaptioning}.

\end{abstract}



\section{Introduction}\label{sec:intro}


Image captioning is a task of mapping images to text for human consumption. Broadly speaking, in order for a caption to be good it must satisfy two requirements: it should be a \emph{fluent}, well-formed phrase or sentence in the target language; and it should be \emph{informative}, or \emph{descriptive}, conveying meaningful non-trivial information about the visual scene it describes. Our goal in the work presented here is to improve captioning on both of these fronts. 

Because these properties are somewhat vaguely defined, objective evaluation of caption quality remains a challenge, more so than evaluation of earlier established tasks like object detection or depth estimation. However, a number of metrics have emerged as preferred, if imperfect, evaluation measures. Comparison to human (``gold standard") captions collected for test images is done by means of metrics borrowed from machine translation, such as BLEU\cite{papineni2002bleu}, as well as new metrics introduced for the captioning task, such as CIDEr\cite{vedantam2015cider} and SPICE\cite{anderson2016spice}.

In contrast, to assess how informative a caption is, we may design an explicitly discriminative task the success of which would depend on how accurately the caption describes the visual input. One approach to this is to consider \emph{referring expressions}~\cite{kazemzadeh2014referitgame}: captions for an image region, produced with the goal of unambiguously identifying the region within the image to the recipient. We can also consider the ability of a recipient to identify an entire image that matches the caption, out of two (or more) images~\cite{vedantam2017context}. This -- caption \emph{discriminability} -- is the focus of our work presented here.

\begin{figure}[t]
\begin{minipage}[t]{.45\linewidth}
\setstretch{.75}
\begin{centering}
\includegraphics[height=2.4cm]{images/COCO_train2014_000000489297.jpg}
\end{centering}
{\footnotesize
{\bf Human}: a large jetliner taking off from an airport runway
\\
{\bf ATTN+CIDER}: a large airplane is flying in the sky\\
{\bf Ours}: a large airplane taking off from \textcolor{green!50!black}{runway}
}
\end{minipage}%
\hspace{2em}
\begin{minipage}[t]{.45\linewidth}
\setstretch{.75}
\begin{centering}
\includegraphics[height=2.4cm]{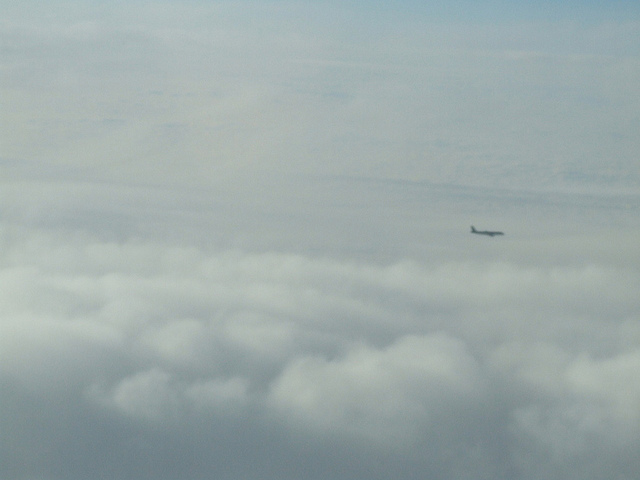}
\end{centering}
{\footnotesize{\bf Human}: a jet airplane flying above the clouds in the distance\\
{\bf ATTN+CIDER}: a large airplane is flying in the sky\\
{\bf Ours}: a plane flying in the sky with a \textcolor{green!50!black}{cloudy sky}
}
\end{minipage}
\caption{Example captions generated by human, an existing automatic system (ATTN+CIDER\cite{rennie2016self}), and a model trained with our proposed method (ATTN+CIDER+DISC(1), see Section~\ref{sec:exp})}
\label{fig:human_vs_machine_vs_baseline} 
\end{figure}

Traditionally used training objectives, such as maximum likelihood estimation (MLE) or CIDEr,
tend to encourage the model to ``play it safe", often yielding overly general captions as illustrated in Fig.~\ref{fig:human_vs_machine_vs_baseline}. Despite the visual differences in the image, a top captioning system~\cite{rennie2016self} produces the same caption for both images. In contrast, humans appear to notice ``interesting" details that are likely to distinguish the image from other potentially similar images, even without explicitly being requested to do so. (We confirm this assertion empirically in Sec.~\ref{sec:exp}.)

To reduce this gap, we propose to incorporate an explicit measure for discriminability into learning a caption generator, as part of the training loss. Our discriminability loss is derived from the ability of a (pre-trained) \emph{retrieval model} to match the correct image to its caption significantly stronger than any other image in a set, and vice versa (caption to correct image above other captions).

Language-based measures like BLEU reward machine captions for mimicking human captions, and so since, as we state above, human captions are discriminative, one could expect these measures to be correlated with descriptiveness. However, in practice, given an imperfect caption generator, there may be a tradeoff between fluency and descriptiveness; our training regime allows us to negotiate this tradeoff and ultimately improve both aspects of a generator. 

Our discriminability loss can be added to any gradient-based learning procedure for caption generators. We show in Sec.~\ref{sec:exp} that it can improve some recently proposed models which currently at or near state of the art for captioning, for all metrics evaluated. In particular, to our knowledge, we establish new state of the art in discriminative captioning.

\section{Related work}\label{sec:related}
\noindent\textbf{Image captioning}
Most modern approaches~\cite{Vinyals2014,Karpathy2015,Ma2016} encode an
image using a convolutional neural network (CNN), and feed this as input to an recurrent network (RNN), typically with some form of gating or memory. The RNN can generate a arbitrary-length sequence of words. Within this generic framework, many efforts~\cite{Vinyals2014,Mao2015, Xu2015,Karpathy2015,lu2016knowing,yang2016review,anderson2017bottom,gu2017stack,wang2017skeleton} explored different encoder-decoder structures, including attention-based models.
There has also been exploration of different training objectives. For example,~\cite{Liu2016,yao2016boosting} add some auxiliary tasks like word appearance prediction;~\cite{wang2017diverse} uses Conditional Variational Autoencoder(CVAE) and optimize over evidence lower bound(ELBO);~\cite{Ranzato2016,rennie2016self,liu2016improved,zhang2017actor,wang2017video,ren2017deep} applied Reinforcement Learning algorithms on captioning, so that the models can be optimized directly on the non-differentiable rewards like CIDEr score, BLEU score etc.

\noindent\textbf{Visual Semantic Embedding methods}
Image-Caption retrieval has been considered as a task relying on image captioning~\cite{Vinyals2014,Karpathy2015,Ma2016,Xu2015}. However, it can also be regarded as a multi-modal embedding task. In previous works~\cite{frome2013devise,Wang2016a,WSABIE,wang2017learning,nam2016dual,Vendrov2015} visual and textual embeddings are trained with the objective to minimize matching loss, e.g., ranking loss on cosine distance, or to enforce partial order on captions and images. 


\noindent\textbf{Discrimination tasks} in the context of caption evaluation were proposed in~\cite{vedantam2017context,Andreas2016}: given a set of other images, called distractors, the generated captions of each image have to distinguish one from others. In the "speaker-listener" model~\cite{Andreas2016}, 
the speaker is trained to generate captions, and a listener to prefer the correct image over a wrong one, given the caption. At test time, the listener re-ranks the captions sampled from the speaker.
\cite{vedantam2017context} propose a decoding mechanism which can suppress the caption elements that are common for both target image and disctractor image. In contrast to our work, both~\cite{vedantam2017context} and~\cite{Andreas2016}  require the distractor to be presented prior to caption production. We aim to generate distinctive captions a-priori, without a specific distractor at hand, like humans appear to do.

Referring expressions is another flavor of discriminative captioning task that has attracted interest after the release of the standard datasets~\cite{kazemzadeh2014referitgame,Yu2016,Mao2016}. \cite{yu2016joint,luo2017comprehension} learned to generate more discriminative referring expressions guided by a referring expression comprehension model. The techniques in those papers are strongly tied to the task of describing a region within an image, while our goal here is to describes natural scenes in their entirety.

\noindent\textbf{Visual Dialog} 
has recently attracted interests in the field~\cite{geman2015visual,das2016visual}.
While it's hard to evaulate generic `chat',
\cite{das2017learning,de2016guesswhat} propose goal-driven visual dialog tasks and datasets. \cite{das2017learning} proposes the `image guessing' game where two agents -- Q-BOT and A-BOT -- who communicate in natural language dialog so that Q-BOT can select an unseen image from a lineup of images.  GuessWhat Game~\cite{de2016guesswhat} is similar, but guess an object in a image during a dialog. In another related effort~\cite{li2017learning} the machine must show understanding the difference between two images by asking a question that has different answers for two images. Our work shares the ultimate purpose (producing text that allows image identification) with these efforts, but in contrast to those, our aim is to generate a caption in a single ``shot". This is somewhat similar to round 0 of the dialog in~\cite{das2017learning}, where the agent is given a caption generated by~\cite{Karpathy2015} (without regard to any discrimination task) and chooses an image from a set. Since our captions are shown in Sec.~\ref{sec:exp} to be both fluent and discriminative, switching to using them may improve/shorten visual dialog.



\noindent\textbf{Similar work} Finally, some recent work is similar to ours in its goals (learning to produce discriminative captions) and, to a degree, in techniques.
The motivation in~\cite{dai2017contrastive} is similar, but the focus is on caption (rather than image) retrieval. The objective is contrastive: pushing the negative captions from different images to have lower probability than positive captions using noise contrastive learning.
In \cite{lu2017best}, more meaningful visual dialog responses are generated by distilling knowledge from a discriminative model trained to rank different dialog responses given the previous dialog context.
\cite{dai2017towards,shetty2017speaking} proposes using Conditional Gernerative Adversarial Network to train image captioning. They both learn a discriminator to distinguish human captions from machine captions. For more detailed discussion of \cite{dai2017contrastive,dai2017towards,shetty2017speaking}, see \supporapp.

Despite being motivated by a desire to improve caption discriminability, all these methods are fundamentally remain tied to the objective of matching the surface form of human captions, and do not include an explicitly discriminative objective in training. 
Ours is the first work incorporate both image retrieval and caption retrieval into caption generation training. We can easily ``plug" our method into existing models, for instance combine it with CIDEr optimization, leading to improvements in metrics across the board: both the discriminative metrics (image identification) and traditional metrics such as ROUGE end METEOR (Tables~\ref{tab:val_result},\ref{tab:disc_result_on_test}).

\section{Models}\label{sec:models}
Our model involves two main ingredients: a \emph{retrieval} model that scores images caption pairs, and a caption generator that maps an image to a caption. We describe the models used in our experiments below; however we note that our approach is very modular, and can be applied to different retrieval models and/or different caption generators.
Then we describe the key element of our approach: combining these two ingredients in a collaborative framework. We use the retrieval score derived from the retrieval model to help guide training of the generator. 

\subsection{Retrieval model}
The retrieval model we use is taken from~\cite{faghri2017vse++}. It is a embedding network which embeds both text and image into a shared semantic space in which a similarity (compatibility) score can be calculated between a caption and an image. We outline the model below, for details see~\cite{faghri2017vse++}.

We start with an image $I$ and caption $c$. First, domain-specific encoders compute an image feature vector $\bfgreek{phi}(I)$, e.g., using a CNN, and a caption feature vector $\bfgreek{psi}(c)$, e.g. using an RNN-based text encoder. These feature vectors are then projected into a joint space by $W_I$ and $W_c$.
\begin{align}
f(i) = \mathbf{W}_I^T\bfgreek{phi}(I)\\
g(c) = \mathbf{W}_c^T\bfgreek{psi}(c)
\end{align}
The similarity score between $I$ and $c$ is now computed as the cosine similarity in the embedding space:
\begin{align}
s(I,c) = \frac{f(I)\cdot g(c)}
{\|f(I)\|\|g(c)\|}
\end{align}
The parameters of the caption embedding $\bfgreek{psi}$, as well as the maps $\mathbf{W}_I$ and $\mathbf{W}_c$, are learned jointly, end-to-end, by minimizing the contrastive loss defined below. In our case, the image embedding network $\bfgreek{phi}$ is a pre-trained CNN and the parameters are fixed during training. 


\noindent\textbf{Contrastive loss} is a sum of two hinge losses:
\begin{align}
L_{\textsc{CON}}(c,I) &=
\max_{c^\prime}[\alpha+s(I,c^\prime)-s(I,c)]_+ \nonumber \\
&+\max_{I^\prime}[\alpha+s(I^\prime,c)-s(I,c)]_+
\label{eq:con_loss}
\end{align}
where $[x]_+ \equiv \max(x,0)$. The max in~\eqref{eq:con_loss} is taken, in practice, over a batch of $B$ images and corresponding captions. The (image,caption) pairs $(I,c)$ are  correct matches, while $(I',c)$ and $(I,c')$ are incorrect (e.g., $c'$ is a caption that does not describe $I$).
Intuitively, this loss ``wants'' the model to assign the matching pair $(I,c)$ the score higher (by at least $\alpha$) than the score of any mismatching pair, either $(I',c)$ or $(I,c')$ that can be formed from the batch. This objective can be viewed as a hard negative mining version of triplet loss~\cite{schroff2015facenet}.





\subsection{Discriminability loss}
The ideal way to measure discriminability is to pass it to human and get feedback from them, like in~\cite{LingNIPS2017}. However it is rather costly and very slow to collect. Here, we propose instead to use a pre-trained retrieval model to work as a proxy for human perception. Specifically, we define the discriminability loss follows.

Suppose we have a captioning system, parameterized by a set of parameters $\bfgreek{theta}$, that can output conditional distribution over captions for an image, $p_c(c|I;\bfgreek{theta})$. Then, the objective of minimizing the discriminability loss is 
\begin{align}
\min_{\bfgreek{theta}}\mathbb{E}_{\widehat{c}\;\sim\;p(c|I;\bfgreek{theta})} \left[L_{\textsc{CON}}(\widehat{c}, I)\right]
\end{align}
In other words, the objective involves the same contrastive loss used to train the retrieval model. However, when training the retrieval model, the loss relies on ground truth image-caption pairs (with human-produced captions), and is back-propagated to update parameters of the retrieval model. 
Now, when using the loss to train caption generators, an input batch (over which the max in~\eqref{eq:con_loss} is computed) will include pairs of images with captions that are sampled from the posterior distribution produced by a caption generator; the signal derived from the loss will be used to update parameters $\bfgreek{theta}$ of the generator, while holding the retrieval model fixed.

\subsection{Caption generation models}
We now briefly describe two caption generation models used in our experiments; both are introduced in~\cite{rennie2016self} where further details can be found. Discussion on training these models with discriminability loss is deferred until Sec.~\ref{sec:training}.

\noindent\textbf{FC Model}
The first model is a simple sequence encoder initialized with visual features. Words are represented with an embedding matrix (a vector per word). Visual features are extracted from an image using a CNN. 

The caption sequence is generated by a form of LSTM model. Its output at time $t$ depends on the previously generated word and on the context/hidden state (evolving as per LSTM update rules). At training time the word fed to the state $t$ is the ground truth word $w_{t-1}$; at test time, it is the predicted word $\widehat{w}_{t-1}$. The first word is a special \textup{BOS} (beginning of sentence) token. The sequence production is terminated when the special \textup{EOS} token is output. The image features (mapped to the dimensions of word embeddings) serve at the initial ``word'' $w_{-1}$, fed to the state at $t=0$.

\noindent\textbf{ATTN model}
The main difference between the second model and the FC model is that each image is now encoded into a set of spacial features: each encodes a sub-region of the image. At each word $t$, the context (and thus the output) depends not only on the previous output and the internal state of the LSTM, but also a weighted average of all the spatial features. This weighted averaging of features is called attention mechanism, and the attention weights are computed by a parametric function.


Both models provide us with a posterior distribution over sequence of words $c=(w_0,\ldots,w_T)$, factorized as
\begin{equation}
p(c|I;\theta)=\prod_t p(w_t|w_{t-1},I;\bfgreek{theta})
\label{eq:p_w}
\end{equation}

\section{Learning to reward discriminability}\label{sec:training}
Given a generator model, we may want to train it to minimize the discriminability loss~\eqref{eq:con_loss}. A natural approach would be to use gradient descent. Unfortunately, the loss is non-differentiable since it involves sampling captions for input images in a batch. 

One way to tackle this is by the Gumbel-softmax reparametrization trick~\cite{jang2016categorical,maddison2016concrete} which has been used in image captioning and visual dialog~\cite{shetty2017speaking,lu2017best}. Instead, in this paper, we follow the philosophy of~\cite{Ranzato2016, hendricks2016generating,yu2016joint,rennie2016self} and treat captioning as a reinforcement learning problem. Specifically we use the REINFORCE algorithm~\cite{williams1992simple}.  In similar contexts, REINFORCE has been applied in~\cite{yu2016joint,hendricks2016generating} to train sequence prediction. Here we use a variant of ``REINFORCE with baseline'' algorithm proposed in the ``self-critical'' approach of~\cite{rennie2016self}, as outlined below.



The objective is to learn parameters $\bfgreek{theta}$ of the policy (here defining a mapping from $I$ to $c$, i.e., $p$) that would maximize the reward computed by function $R(c,I)$. The algorithm computes an update to approximate the gradient of the expected reward (a function of stochastic policy parameters), known as the policy gradient:
\begin{equation}
\nabla_{\bfgreek{theta}} E_{\widehat{c}\sim p(c|I;\bfgreek{theta})}[R(\widehat{c},I)] \approx (R(\widehat{c},I) - b) \nabla_{\bfgreek{theta}} \log p(\widehat{c}|I;\bfgreek{theta})
\label{eq:policy-gradient}
\end{equation}

Here $\widehat c$ represents the caption sampled from~\eqref{eq:p_w}.
The \emph{baseline} $b$ is computed by a function designed to make it independent of the sample (leading to variance reduction without increasing bias~\cite{weaver2001optimal}). In our case, folllowing~\cite{rennie2016self}, the baseline is the value of the reward $R(c^\ast, I)$ on the greedy decoding\footnote{We also tried setting $b$ to the reward of ground truth caption, and found no significant difference.} output $c^\ast=(\textup{BOS},w^\ast_1,\ldots,w^\ast_T)$,

\[
w^*_t = \argmax_w p(w|w^*_{0,\ldots,t-1},I)
\]

We could apply this to maximizing the reward defined simply as the negative discriminability loss $-L_{\textsc{CON}}(\widehat{c}, I)$. However, as observed in previous work~\cite{luo2017comprehension}, this does not yield human-friendly captions since discriminability loss will not directly hurt from influency. So we will combine the discriminability loss with other, traditional objectives in defining the reward, as described below.

\subsection{Training with maximum likelihood}
The standard objective in training a sequence prediction model is to maximize word-level log-likelihood, which for a pair $(I,c)$ is defined as $R_{\textsc{LL}}(c,I)=\log p(c|I;\bfgreek{theta})$.
The parameters $\bfgreek{theta}$ here include word embedding matrix and LSTM weights which are updated as part of training, and the CNN weights, which are held fixed after pre-training on a vision task such as ImageNet classification.
This reward can be directly maximized via gradient ascent (equivalent to gradient descent on the cross-entropy loss), yielding maximum likelihood estimate (MLE) of the model parameters.

Combining the log-likelihood reward with discriminability loss in the REINFORCE framework corresponds to defining the reward as
\begin{equation}
R(c,I)=R_{\textsc{LL}}(c,I)\,-\,\lambda L_{\textsc{CON}}(\widehat c,I),
\label{eq:R-ll}
\end{equation}
yielding the policy gradient:
\begin{equation}
\begin{split}
\nabla_{\bfgreek{theta}}&E[R(c,I)]\approx\nabla_{\bfgreek{theta}}R_{\textsc{LL}}(c,I)\\
&-\,\lambda
\left[L_{\textsc{CON}}(\widehat{c},I)-L_{\textsc{CON}}(c^\ast,I)\right]
\nabla_{\bfgreek{theta}}\log p(\widehat{c}|I;\bfgreek{theta})
\end{split}
\label{eq:mle-policy}
\end{equation}
The coefficient $\lambda$ determines the tradeoff between matching human captions (expressed by the cross-entropy) and discriminative properties expressed by $L_{\textup{CON}}$. 

\subsection{Training with CIDEr optimization}
In our experiments, it was hard to train with the combined objective in~\eqref{eq:mle-policy}. For small $\lambda$, the solutions seemed stuck in a local minimum; but increasing $\lambda$ would abruptly make output less fluent.

An alternative to MLE is to train the model to maximize some other reward/score, such as BLEU or METEOR. Here if pursue optimization of the CIDEr score~\cite{vedantam2015cider}. CIDEr measures consensus in image captions by performing a TF-IDF weighting for each n-gram, and optimizing over CIDEr can also benefit other metrics\cite{rennie2016self}.
We found that in practice, the discriminability loss appears to ``cooperate" better with CIDEr than with log-likelihood; we also observed better performance, across many metrics, on validation set as described in Section~\ref{sec:exp}. 


Compared to \cite{rennie2016self}, which uses CIDEr as reward function, the difference here is we use a weighted sum of cider score and discriminability loss.
\begin{equation}
\nabla_\theta E[R(\widehat{c},I)] \approx (R(\widehat{c},I) - R(c^\ast, I)) \nabla_\theta \log p(\hat c|I;\bfgreek{theta}),
\end{equation}
where the reward is the combination
\begin{equation}
R(\widehat{c},I) = \mbox{CIDEr}(\widehat{c}) - \lambda L_{\textsc{CON}}(\widehat{c}, I),
\end{equation}
with $\lambda$ again representing the relative weight of discriminability loss vs. CIDEr.





\section{Experiments and results}\label{sec:exp}
The main goal of our experiments is to evaluate the utility of the proposed discriminability objective in training image captions. Recall that our motivation for introducing this objective is two-fold: to make the captions more discriminative, and to improve caption quality in general (with the implied assumption that expected discriminability is part of the unobservable human ``objective" in describing images).

\noindent\textbf{Dataset.} We train and evaluate our model on COCO dataset \cite{lin2014microsoft}. To enable direct comparisons, we use the data split from \cite{vedantam2017context}, which includes 113,287 images for training, 5,000 images for validation, and another 5,000 held out for test. Each image is associated with five human captions.

\subsection{Implementation details}\label{sec:details}

As the basis for caption generators, we used two models described in Section~\ref{sec:models}, FC and ATTN, with relevant implementation details as follows.

For image encoder in retrieval and FC captioning model, we used a pretrained Resnet-101~\cite{renNIPS15fasterrcnn}. For each image, we take the global average pooling of the final convolutional layer output, which results in a vector of dimension 2048.
The spatial features are extracted from output of a Faster R-CNN\cite{renNIPS15fasterrcnn,anderson2017bottom} with ResNet-101\cite{he2016deep}, trained by object and attribute annotations from Visual Genome\cite{krishna2017visual}. The number of spatial features varies from image to image. Each feature encodes a region in the image which is proposed by region proposal network. Both the FC features and Spatial features are pre-extracted, and no finetuning is applied on image encoders.
For captioning models, the dimension of LSTM hidden state, image feature embedding, and word embedding are all set to 512. 

The retrieval model uses GRU-RNN to encode text, and the FC features above as the image feature. The word embedding has 300 dimensions and the GRU hidden state size and joint embedding size are 1024. The margin $\alpha$ is set to 0.2, as suggested by~\cite{faghri2017vse++}.


\noindent\textbf{Training}
All of our captioning models are trained according to the following scheme. We first pretrain the captioning model using MLE, with Adam\cite{kingma2014adam}. After 40 epochs, the model is switched to self-critical training with appropriate reward (CIDEr alone or CIDEr combined with discriminability)  and continue for another 20 epochs. For fair comparison, we also train another 20 epochs for MLE-only models.

For both retrieval and captioning models, the batch size is set to 128 images. The learning rate is initialized to be 5e-4 and decay by a factor 0.8 for every three epochs.

During test time, we apply beam search to sample captions from captioning model. The beam size is set to 2.

\subsection{Experiment design}
We consider a variety of possible combination of captioning objective (MLE/CIDEr), captioning model (FC/ATTN), and inclusion/exclusion of discriminability, abbreviating the model references for brevity, so, e.g., ATTN+CIDER+DISC(5) corresponds to fine-tuning the attention model with a combination of CIDEr and discriminability loss, with $\lambda=5$.

\noindent\textbf{Evaluation metrics} Our experiments consider two families of metrics. The first family of standard metrics that have been proposed for caption evaluation, mostly based on comparing generated captions to human ones, includes BLEU\cite{papineni2002bleu}, METEOR\cite{denkowski2014meteor}, ROUGE\cite{lin2004rouge}, CIDEr\cite{vedantam2015cider} and SPICE\cite{anderson2016spice}.

The second set of metrics directly assesses how discriminative the captions are. This includes automatic assessment, by measuring accuracy of the trained retrieval model on generated captions.

We also assess how discriminative the generated captions are when presented to humans. To measure this, we conducted an image discrimination task on Amazon Mechanical Turk (AMT), following the protocol in~\cite{vedantam2017context}.
A single task (HIT) involves displaying, along with a caption, a pair of images (in randomized order). One image is the \emph{target} for which the caption was generated, and the second is a \emph{distractor} image. The worker is asked to select which image is more likely to match the caption. Each target/distractor pair is presented to five distinct workers; we report the fraction of HITs with correct selection by at least $k$ out of five workers, with $k=3,4,5$. Note that $k=3$ suffers from highest variance since the forced choice nature of the task would produce non-trivial chance of 3/5 correct selections when the caption is random. In our opinion, $k=4$ is the most reliable indicator of human ability to discriminate based on the caption.

The test set used for this evaluation is the set from \cite{vedantam2017context}, constructed as follows. For each image in the original test set, its nearest neighbor is found based on visual similarity, estimated as Euclidean distance between the FC7 feature vectors computed by VGG-16 pre-trained on ImageNet~\cite{simonyan2014very}. Then a captioning model is run on the nearest neighbor images, and the word-level overlap (intersection over union) of the generated captions is used to and pick (out of 5000 pairs) the top (highest overlap) 1000 pairs.

For preliminary evaluation, we followed a similar protocol to construct our own validation set of target/distractor pairs; both the target images and distractor images were taken from the caption validation set (and so were never seen by any training procedure).

\subsection{Retrieval model quality}

Before proceeding with main experiments, we report in Tab.~\ref{tab:retrieval-result} the accuracy of the retrieval model on validation set, with human-generated captions. This is relevant since we rely on this model as a proxy for discriminability in our training procedure. While this model does not achieve state of the art for image caption retrieval, it is good enough for providing training signal to improve caption results.

{\renewcommand{\arraystretch}{0.8}
\begin{table}[!tbh]\small
  \centering
    \begin{tabular}{l|rrrrr}
          & \multicolumn{1}{l}{R@ 1} & \multicolumn{1}{l}{R@ 5} & \multicolumn{1}{l}{R@ 10} & \multicolumn{1}{l}{Med r} & \multicolumn{1}{l}{Mean r} \\
    \midrule
          & \multicolumn{5}{c}{Caption Retrieval} \\
    \midrule
    1k val & 63.9  & 90.4  & 95.9  & 1.0   & 2.9 \\
    5k val & 38.0  & 68.9  & 81.1  & 2.0   & 10.4 \\
    \midrule
          & \multicolumn{5}{c}{Image Retrieval} \\
    \midrule
    1k val & 47.9  & 80.7  & 89.9  & 2.0   & 7.7 \\
    5k val & 26.1  & 54.7  & 67.5  & 4.0   & 34.6 \\
    \end{tabular}
  \caption{Retrieval model performance on validation set.}\label{tab:retrieval-result}
\end{table}
}

\begin{table*}[!th]\small
  \centering
    \begin{tabular}{lcccccccccc}
          & Acc  & Acc-new  & BLEU4 & METEOR & ROUGE & CIDEr & SPICE & 3 in 5 & 4 in 5 & 5 in 5 \\
    \midrule
    FC+MLE\cite{rennie2016self} & 77.23\% & 77.23\% & \textbf{0.3308} & 0.2566 & 0.5407 & 1.0005 & 0.1855 & 71.78\% & 50.28\% & 18.79\% \\
    FC+CIDER\cite{rennie2016self} & 74.00\% & 74.32\% & 0.3249 & 0.2550 & 0.5428 & 1.0154 & 0.1899 & 73.04\% & 50.58\% & 24.83\% \\
    FC+MSE+DISC (100) & 87.42\%	& 87.42\% &	0.2902 & 0.2523 & 0.5261 & 0.9190 & 0.1881 & 76.91\% & 54.62\% & 23.20\% \\
    FC+CIDER+DISC (1) & 79.26\% & 79.49\% & 0.3274 & \textbf{0.2574} & \textbf{0.5457} & \textbf{1.0231} & \textbf{0.1939} & 74.26\% & 55.53\% & 24.13\% \\
    FC+CIDER+DISC (5) & 85.90\% & 85.68\% & 0.3072 & 0.2534 & 0.5382 & 0.9678 & 0.1904 & 78.63\% & 58.03\% & 32.64\% \\
    FC+CIDER+DISC (10) & \textbf{88.69\%} & \textbf{88.01\%} & 0.2727 & 0.2473 & 0.5224 & 0.8795 & 0.1807 & \textbf{80.01\%} & \textbf{62.71\%} & \textbf{37.15\%} \\
    \midrule
    ATTN+MLE\cite{rennie2016self} & 72.40\% & 73.12\% & 0.3582 & 0.2719 & 0.5649 & 1.1078 & 0.2019 & 69.90\% & 54.60\% & 28.07\% \\
    ATTN+CIDER\cite{rennie2016self} & 71.05\% & 71.13\% & 0.3592 & 0.2695 & 0.5678 & 1.1332 & 0.2083 & 69.97\% & 51.34\% & 27.34\% \\
    ATTN+MLE+DISC (100) & 82.64\% & 83.03\% & 0.3266 & 0.2697 & 0.5542 & 1.0448 & 0.2057 & 78.18\% & 55.63\% & 21.71\% \\
    ATTN+CIDER+DISC (1) & 75.74\% & 76.60\% & \textbf{0.3627} & \textbf{0.2728} & \textbf{0.5706} & \textbf{1.1406} & \textbf{0.2113} & 72.70\% & 53.23\% & 34.33\% \\
    ATTN+CIDER+DISC (5) & 80.98\% & 81.43\% & 0.3504 & 0.2704 & 0.5636 & 1.1026 & 0.2097 & 76.69\% & 60.94\% & 33.49\% \\
    ATTN+CIDER+DISC (10) & \textbf{83.69\%} & \textbf{83.50\%} & 0.3261 & 0.2673 & 0.5549 & 1.0552 & 0.2070 & \textbf{81.93\%} & \textbf{65.12\%} & \textbf{35.41\%} \\
    \bottomrule
    \end{tabular}%
  \caption{Automatic scores and human-study discriminability on the validation set. The numbers in the parenthesis are discriminability loss weight $\lambda$.}\label{tab:val_result}
\end{table*}%

\subsection{Captioning performance}

In Table \ref{tab:val_result}, we show the results on validation set with a variety of model/loss settings. Note that all the FC*/ATTN* model with different settings are finetuned from the same model pre-trained with MLE. 
The results in the table for the machine scores are based on all the 5k images. For discriminability, we randomly select a subset of 300 image pairs from validation set.
We can draw a number of conclusions from these results.

\noindent\textit{Effectiveness of reinforcement learning.} In the first column, we report the retrieval accuracy (Acc, \% of pairs in which the model correctly selects the target vs. distractor) on pairs given the output of the captioning model. Training with the discriminability loss produces higher values here, meaning that our captions are more discriminative \emph{to the retrieval model}, as intended. As a control experiment, we also report the accuracy (Acc-new) obtained by a same architecture but separately trained retrieval model, not used in training caption generators. Acc and Acc-new are very similar for all models, showing that our model does not overfit to the retrieval model it uses during training time.

\noindent\textit{Human discrimination.} More importantly, we observe that incorporating discriminability in training yields captions that are more discriminative to humans, with higher $\lambda$ leading to better human accuracy. 

\noindent\textit{Improved caption quality.} We also see that, as hoped, incorporating discriminability indeed improves caption quality as measured by a range of metrics that are not tied to discrimination, such as BLEU etc. Even the CIDEr scores are improved when adding discriminability to the CIDEr optimization objective with moderate $\lambda$. This is somewhat surprising since the addition of $L_{\textsc{CON}}$ could be expected to detract from the original objective of maximizing CIDEr; we presume that the improvement is due to the additional objective ``nudging" the RL process and helping it escape less optimal solutions.

\noindent\textit{Model/loss selection} While discriminability loss works for both ATTN model and FC model, and with both MLE and CIDEr learning, to make captions more discriminative, and with mild $\lambda$ to improve other metrics, the overall performance analysis favors ATTN+CIDEr combination. We also note that ATTN is better than FC on discriminability metrics even when trained without $L_{\textup{CON}}$, but the gains are less significant than in automatic metrics.

\noindent\textit{Effect of $\lambda$} As stated above, mild $\lambda=1$, combined with ATTN+CIDER, appear to yield the optimal tradeoff, improving measures of discriminative \emph{and} descriptive quality across the board. Higher values of $\lambda$ do make resulting captions more discriminative to both humans and machines, but at the cost of reduction in other metrics, and in our observations (see Section~\ref{sec:qual}) in perceived fluency. This analysis is applicable across model/loss combinations.
We also notice a relative large range of $\lambda$(0.5-1.2) can yield similar improvement on automatic metrics.


\begin{table*}[ht]\small
  \centering
    \begin{tabular}{lcccccccccc}
          & Acc  & Acc-new & BLEU4 & METEOR & ROUGE & CIDEr & SPICE & 3 in 5 & 4 in 5 & 5 in 5 \\
    \midrule
    Human & 74.30\% & 74.14\% & -     & -     & -     & -     & -     & 91.14\% & 82.38\% & 57.08\% \\
    \midrule
    ATTN+MLE\cite{rennie2016self} & 68.60\% & 66.90\% & 0.3907 & 0.2913 & 0.5956 & 1.2198 & 0.2132 & 72.06\% & 59.06\% & 44.25\% \\
    ATTN+CIDER\cite{rennie2016self} & 68.19\% & 65.12\% & 0.3871 & 0.2908 & 0.5971 & 1.2604 & 0.2260 & 70.07\% & 55.95\% & 35.95\% \\
    \midrule
    CACA\cite{vedantam2017context} & 75.80\% & 76.00\% & 0.2357 & 0.2186 & 0.4719 & 0.7656 & 0.1526 & 74.1\%\footnotemark[1] & 56.88\%\footnotemark[1] & 35.19\%\footnotemark[1] \\
    ATTN+CIDER+DISC(1) & 72.63\% & 70.68\% & \textbf{0.3971} & \textbf{0.2931} & \textbf{0.6043} & \textbf{1.2770} & \textbf{0.2302} & 76.91\% & 61.67\% & 40.09\% \\
    ATTN+CIDER+DISC(10) & \textbf{79.75\%} & \textbf{79.14\%} & 0.3538 & 0.2821 & 0.5811 & 1.1429 & 0.2204 & \textbf{77.70\%} & \textbf{64.63\%} & \textbf{44.63\%} \\
    \end{tabular}%
  \caption{Automatic scores and discriminability on 1k test set.}\label{tab:disc_result_on_test}%
\end{table*}%

Following the observations above, we select a subset of methods to evaluate on the (previously untouched) test set, with results shown in Table~\ref{tab:disc_result_on_test}.

Here, we add two more results for comparison. The first involves presenting AMT workers with \emph{human} captions for the target images. Recall that these captions are collected for each image independently, without explicit instructions related to discriminability, and without showing potential distractors. However, human captions prove to be highly discriminative. This, not surprisingly, indicates that humans may be incorporating an implicit objective of describing elements in an image that are surprising, notable or otherwise may help in distinguishing the scene from other, scenes. While this performance is not perfect (4/5 accuracy of 82\%) it is much higher than for any automatic caption model.

The second additional set of results is for the model in~\cite{vedantam2017context}, evaluated on captions provided by the authors. Note that in contrast to our model (and to human captions), this method has the benefit of seeing the distractor prior to generating the caption; nonetheless, its performance is dominated across metrics by our attention models trained with CIDEr optimization combined with discriminability loss. It also appears that this models'  gains on discrimination are offset by a significant deterioration under other metrics.

In contrast, our ATTN+CIDER+DISC model with $\lambda=10$ achieves the most discriminative image captioning result without major degradation under other metrics; the ATTN+CIDER+DISC with $\lambda=1$ again shows the best discriminability/descriptiveness tradeoff among the evaluated models.

\begin{figure}[!th]
\begin{minipage}[t]{.45\linewidth}
\setstretch{.7}
\includegraphics[height=2.4cm]{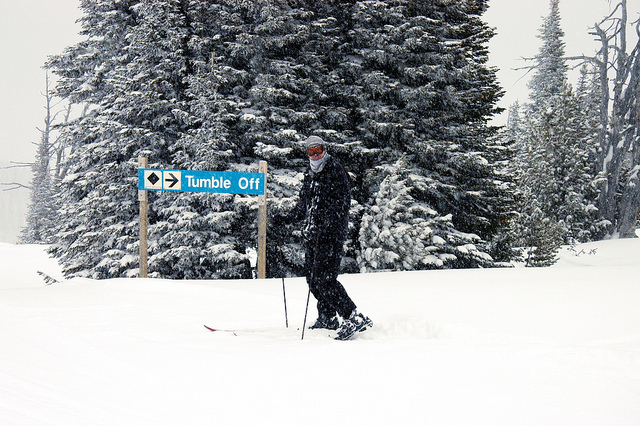}\\
{\footnotesize
{\bf Human}: a man riding skis next to a blue sign near a forest
\\
{\bf ATTN+CIDER}: a man standing on skis in the snow\\
{\bf Ours}: a man standing in the snow with \textcolor{green!50!black}{a sign}
}
\end{minipage}%
\hspace{1em}
\begin{minipage}[t]{.45\linewidth}
\setstretch{.7}
\includegraphics[height=2.4cm]{images/COCO_val2014_000000464144.jpg}\\
{\footnotesize
{\bf Human}: the man is skiing down the hill with his goggles up\\
{\bf ATTN+CIDER}: a man standing on skis in the snow\\
{\bf Ours}: a man riding skis on a snow covered slope
}
\end{minipage}\\
\begin{minipage}[t]{.45\linewidth}
\setstretch{.7}
\includegraphics[height=2.4cm]{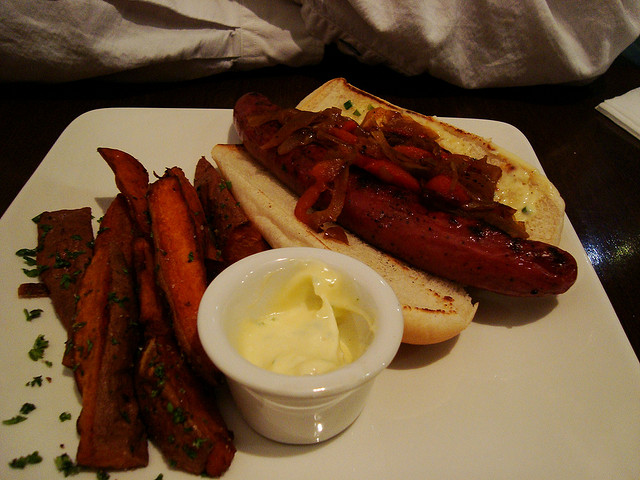}\\
{\footnotesize
{\bf Human}: a hot dog serves with fries and dip on the side
\\
{\bf ATTN+CIDER}: a plate of food with meat and vegetables on a table\\
{\bf Ours}: \textcolor{green!50!black}{a hot dog and french fries} on a plate
}
\end{minipage}%
\hspace{1em}
\begin{minipage}[t]{.45\linewidth}
\setstretch{.7}
\includegraphics[height=2.4cm]{images/COCO_train2014_000000337365.jpg}\\
{\footnotesize
{\bf Human}: a plate topped with meat and vegetables and sauce\\
{\bf ATTN+CIDER}: a plate of food with meat and vegetables on a table\\
{\bf Ours}: a plate of food with \textcolor{green!50!black}{carrots and vegetables} on a plate
}
\end{minipage}
\\
\begin{minipage}[t]{.45\linewidth}
\setstretch{.7}
\includegraphics[height=2.4cm]{images/COCO_train2014_000000551470.jpg}\\
{\footnotesize
{\bf Human}: a train on an overpass with people under it
\\
{\bf ATTN+CIDER}: a train is on the tracks at a train station\\
{\bf Ours}: a \textcolor{green!50!black}{red} train parked on the side of a building
}
\end{minipage}%
\hspace{1em}
\begin{minipage}[t]{.45\linewidth}
\setstretch{.7}
\includegraphics[height=2.4cm]{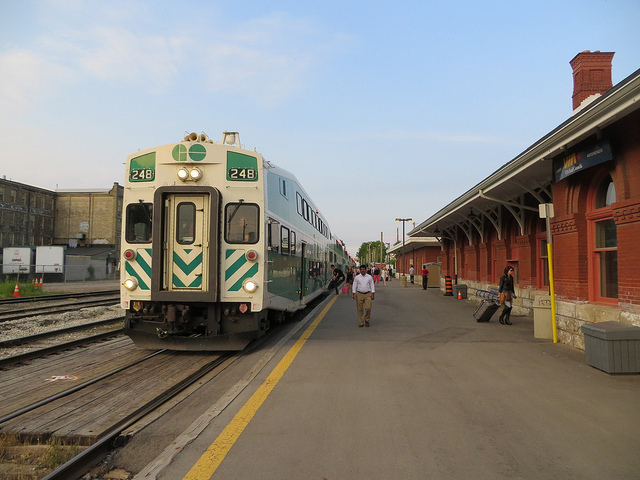}\\
{\footnotesize
{\bf Human}: a train coming into the train station\\
{\bf ATTN+CIDER}: a train is on the tracks at a train station\\
{\bf Ours}: a \textcolor{green!50!black}{green} train traveling down a train station
}
\end{minipage}
\caption{Examples of image captions; Ours refers to ATTN+CIDER+DISC(1)}\label{fig:qual}
\end{figure}

\begin{figure}

\begin{minipage}[t]{.30\linewidth}
\includegraphics[height=2.5cm]{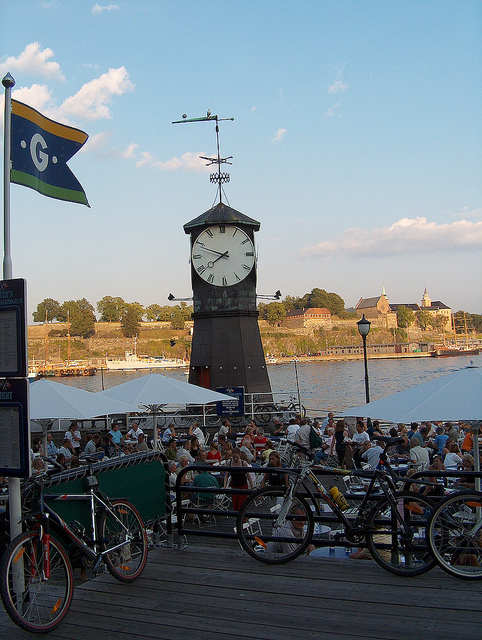}
\end{minipage}%
\hspace{1em}
\begin{minipage}[t]{.60\linewidth}
\includegraphics[height=2.5cm]{images/COCO_val2014_000000527248.jpg}
\end{minipage}\\
\setstretch{.7}
{\footnotesize
{\bf ATTN+MLE}: a large clock tower with a clock on it
\\
{\bf ATTN+CIDER}: a clock tower with a clock on the side of it\\
{\bf ATTN+CIDER+DISC(1)}: a clock tower with \textcolor{green!50!black}{bikes} on the side of \textcolor{green!50!black}{a river}\\
{\bf ATTN+CIDER+DISC(10)}: a clock tower with \textcolor{green!50!black}{bicycles} on the \textcolor{green!50!black}{boardwalk} near \textcolor{green!50!black}{a harbor}
}

\begin{minipage}[t]{\linewidth}
\begin{minipage}[t]{.35\linewidth}
\includegraphics[height=2.5cm]{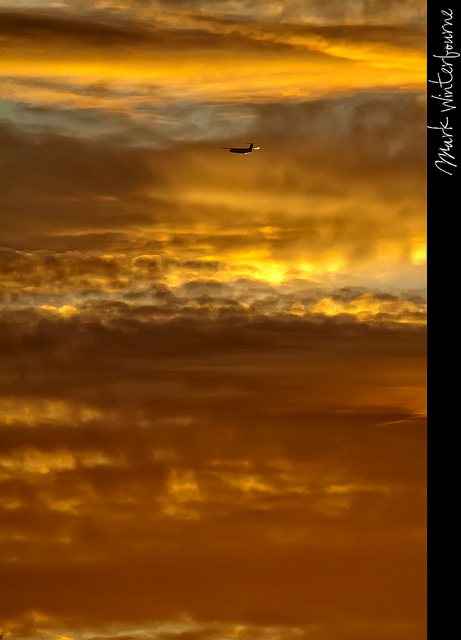}
\end{minipage}%
\hspace{1em}
\begin{minipage}[t]{.55\linewidth}
\includegraphics[height=2.5cm]{images/COCO_val2014_000000345329.jpg}
\end{minipage}
\end{minipage}
\setstretch{.75}
{\footnotesize
{\bf ATTN+MLE}: a view of an airplane flying through the sky
\\
{\bf ATTN+CIDER}: a plane is flying in the sky\\
{\bf ATTN+CIDER+DISC(1)}: a plane flying in the sky with \textcolor{green!50!black}{a sunset}\\
{\bf ATTN+CIDER+DISC(10)}: \textcolor{green!50!black}{a sunset} of \textcolor{red}{a sunset with a sunset in the sunset}
}


\begin{minipage}[t]{\linewidth}
\begin{minipage}[t]{.45\linewidth}
\includegraphics[height=2.5cm]{images/COCO_train2014_000000334400.jpg}
\end{minipage}%
\hspace{3em}
\begin{minipage}[t]{.45\linewidth}
\includegraphics[height=2.5cm]{images/COCO_val2014_000000293782.jpg}
\end{minipage}
\end{minipage}
\setstretch{.75}
{\footnotesize
{\bf ATTN+MLE}: a couple of people standing next to a stop sign
\\
{\bf ATTN+CIDER}: a stop sign on the side of a street\\
{\bf ATTN+CIDER+DISC(1)}: a stop sign in front of \textcolor{green!50!black}{a store with umbrellas}\\
{\bf ATTN+CIDER+DISC(10)}: a stop sign sitting in front of \textcolor{green!50!black}{a store} with \textcolor{red}{shops}
}

\caption{Captions from different models describing the target images(left). Right images are the corresponding distractors selected in val/test set; these pairs were included in AMT experiments.}
\label{fig:amt}
\end{figure}

\textbf{Effect on SPICE score}
To further break down how our discriminabilty loss does, we analyze the affect of different models on the SPICE score~\cite{vedantam2015cider}. It estimates caption quality by transforming both candidate and reference (human) captions into a scene graph
and computing the matching between the graphs.
SPICE is known to have higher correlation with human ratings than other conventional metrics. Furthermore, it provides subclass scores on Color, Attributes, Cardinality, Object, Relation, Size. We report the results on these (on validation set) in detail for different models in Table~\ref{tab:model_spice}.

By adding the discriminability loss, we improve scores on Color, Attribute, and Cardinality. With the latter, qualitative results suggest that the improvement may be due to a refined ability to distinguish ``one" or ``two" from ``group of" or ``many". With small $\lambda$, we can also get the best score on Object. Since the object score is dominant in SPICE, $\lambda=1$ also obtaines highest SPICE score overall in Tables~\ref{tab:val_result},\ref{tab:disc_result_on_test}.

Finally, we can evaluate the diversity in captions generated by different models. We find that including discriminability objective, and using higher $\lambda$, are correlated with captions that are more diverse (4471 distinct captions with ATTN+CIDER+DISC(10) for the 5000 images in validation set, compared to 2640 with ATTN+CIDER) and slightly longer (avg. length 9.84 with ATTN+CIDER+DISC(10) vs. 9.20 with ATTN+CIDER). Detailed analysis can be found in the 
\supporapp.

\footnotetext[1]{3 in 5 is quoted from~\cite{vedantam2017context}; 4 in 5 and 5 in 5 computed by us on the set of captions provided by the authors.}

{
\setlength{\tabcolsep}{1.8pt}
\begin{table}[ht]\footnotesize
  \centering
    \begin{tabular}{lcccccc}
          & Color & Attribute & Cardinality & Object & Relation & Size \\
    \midrule
    FC+MLE &                9.32  &                8.74  &                1.73  &              34.04  &                4.81  & \textbf{              2.74 } \\
    FC+MLE+D(100) & \textbf{             15.85 } & \textbf{             10.31 } &               5.33  &              34.57  &                4.43  &                2.62  \\
    FC+C  &                5.77  &                7.01  &                1.80  &              35.70  &                5.17  &                1.70  \\
    FC+C+D (1) &                8.28  &                7.81  &                3.45  & \textbf{            36.37 } & \textbf{               5.25 } &               2.10  \\
    FC+C+D (5) &              10.87  &                9.11  &                6.72  &              35.58  &                4.75  &                2.08  \\
    FC+C+D (10) &              12.80  &                9.90  & \textbf{              8.50 } &             34.60  &                4.40  &                1.70  \\
    \midrule
    ATTN+MLE &              11.78  &              10.13  &                3.00  &              36.42  &                5.52  &                3.67  \\
    ATTN+MLE+D(100) & \textbf{            15.80 } & \textbf{             11.83 } &             14.30  &              37.16  &                5.13  & \textbf{              3.97 } \\
    ATTN+C &               7.24  &                8.77  &                8.93  &              38.38  & \textbf{              6.21 } &               2.39  \\
    ATTN+C+D (1) &                9.25  &                9.49  &              10.51  & \textbf{            38.96 } &               5.91  &                2.58  \\
    ATTN+C+D (5) &              11.99  &              10.40  &              15.23  &              38.57  &                5.59  &                2.53  \\
    ATTN+C+D (10) &              12.88  &              10.88  & \textbf{            15.72 } &             38.09  &                5.35  &                2.53  \\
    \bottomrule
    \end{tabular}%
  \caption{SPICE subclass scores on 5k validation set. All the scores here are scaled up by 100. (Here +C means using CIDEr optimzation; +D(x) means using discriminability loss with $\lambda$ being x).}  \label{tab:model_spice}%
\end{table}%
}

\subsection{Qualitative result}\label{sec:qual}
In Figures~\ref{fig:human_vs_machine_vs_baseline},\ref{fig:qual}, we show a sample of validation set images and for each include a human caption, the caption generated by ATTN+CIDER, and our captions produced by ATTN+CIDER+DISC(1). To emphasize the discriminability gap, the images are organized in pairs\footnote{Note that these pairs are formed for the purpose of this figure; these are not pairs shown to AMT workers for human evaluation.}, where both images have the same ATTN+CIDER caption; although these are mostly correct, compared to our and human result, they tend to lack discriminative specificity. 


To illustrate the task on which we base our evaluation of discriminability to humans, we show in Figure~\ref{fig:amt} a sample of image pairs and associated captions. In each case, the target is on the left (in AMT experiments the order was randomized), and we show captions produced by four automtic systems, two without added discriminability objective in training, and two with (with low and high $\lambda$, respectively). Again, we can see that discriminability loss encourages learning to produce more discriminative captions, and that with higher $\lambda$ this may be associated with reduced fluency. We highlight in \textcolor{green!50!black}{green} caption elements that (subjectively) seem to aid discriminability, and in \textcolor{red}{red} the portions that seem incorrect or jarringly non-fluent. For additional experimental results, see \supporapp.
\section{Conclusions}\label{sec:end}
We have demonstrated that incorporating a discriminability loss, derived from the loss of a trained image/caption retrieval model, in training image caption generators improves the quality of resulting captions across a variety of properties and metrics. It does, as expected, lead to captions that are more discriminative, allowing both human recipients and machines to better identify an image being described, and thus arguably conveying more valuable information about the images. More surprisingly, it also yields captions that are scored higher on metrics not directly related to discrimination, such as BLEU/METEOR/ROUGE/CIDEr as well as SPICE, reflecting more descriptive captions. This suggests that richer, more diverse sources of training signal may further improve training of caption generators.


In future work, we plan to explore more sophisticated visual semantic embedding model, which could potentially give better guidance to training than our current retrieval model. We are also interested in how to make it even more discriminative. 


\ifcvprfinal
\paragraph{Acknowledgments} This work was partially supported by NSF award 1409837 and by the ONR award to MIT Lincoln Lab  FA8721-05-C-0002.
\fi

{\small
\bibliographystyle{unsrt}
\bibliographystyle{ieee}
\bibliography{ref}
}

\ifarxiv
\newpage
\appendix
\appendixpage
\section{Result on standard split and MSCOCO test server}
In the paper report results on the split\cite{vedantam2017context}
Here we also report results (with automatic metrics) on the commonly used split introduced in\cite{Karpathy2015} and on COCO test server. Our observations and conclusions are further confirmed by this evaluation.

{
\renewcommand{\arraystretch}{0.7}
\setlength{\tabcolsep}{2pt}
\begin{table}[htbp]\footnotesize
  \centering
    \begin{tabular}{lccccc}
          & BLEU4 & METEOR & ROUGE & CIDER & SPICE \\
    \midrule
          & \multicolumn{5}{c}{val} \\
    \midrule
    ATTN+C & 0.3511 & 0.2700 & 0.5676 & 1.1225 & 0.2043 \\
    ATTN+C+D(1) & \textbf{0.3583} & \textbf{0.2733} & \textbf{0.5720} & \textbf{1.1381} & \textbf{0.2094} \\
    ATTN+C+D(5) & 0.3416 & 0.2698 & 0.5633 & 1.0902 & 0.2054 \\
    ATTN+C+D(10) & 0.3196 & 0.2664 & 0.5520 & 1.0402 & 0.2024 \\
    \midrule
          & \multicolumn{5}{c}{test} \\
    \midrule
    ATTN+C & 0.3566 & 0.2710 & 0.5688 & 1.1345 & 0.2058 \\
    ATTN+C+D(1) & \textbf{0.3614} & \textbf{0.2738} & \textbf{0.5729} & \textbf{1.1425} & \textbf{0.2105} \\
    ATTN+C+D(5) & 0.3439 & 0.2696 & 0.5640 & 1.1018 & 0.2082 \\
    ATTN+C+D(10) & 0.3203 & 0.2671 & 0.5531 & 1.0530 & 0.2050 \\
    \bottomrule
    \end{tabular}%
  \caption{Automatic scores on standard split. (Here +C means using CIDEr optimzation; +D(x) means using discriminability loss with $\lambda$ being x)}
  \label{tab:ka_scores}%
\end{table}%
}

{
\renewcommand{\arraystretch}{0.7}
\setlength{\tabcolsep}{1.8pt}
\begin{table}[htbp]\footnotesize
  \centering
    \begin{tabular}{lcccccc}
          & \multicolumn{1}{c}{Color} & \multicolumn{1}{c}{Attribute} & \multicolumn{1}{c}{Cardinality} & \multicolumn{1}{c}{Object} & \multicolumn{1}{c}{Relation} & \multicolumn{1}{c}{Size} \\
    \midrule
          & \multicolumn{6}{c}{val} \\
    \midrule
    ATTN+C & 6.27 & 8.19 & 9.07 & 38.18 & 5.67 & \textbf{2.76} \\
    ATTN+C+D(1) & 8.17 & 8.89 & 10.94 & \textbf{38.97} & \textbf{5.70} & 2.41 \\
    ATTN+C+D(5) & 10.83 & 9.70 & 13.98 & 38.27 & 5.19 & 2.68 \\
    ATTN+C+D(10) & \textbf{12.33} & \textbf{10.25} & \textbf{14.47} & 37.72 & 4.87 & 2.25 \\
    \midrule
          & \multicolumn{6}{c}{test} \\
    \midrule
    ATTN+C & 5.54 & 7.82 & 7.83 & 38.43 & 6.04 & 2.21 \\
    ATTN+C+D(1) & 7.55 & 8.72 & 9.75 & \textbf{39.08} & \textbf{6.05} & 2.24 \\
    ATTN+C+D(5) & 10.46 & 9.89 & 13.05 & 38.46 & 5.66 & \textbf{2.59} \\
    ATTN+C+D(10) & \textbf{12.21} & \textbf{10.42} & \textbf{14.84} & 38.04 & 5.19 & 2.54 \\
    \bottomrule
    \end{tabular}%
  \caption{SPICE subclass scores on standard split. All the scores here are scaled up by 100.}
  \label{tab:ka_spice}%
\end{table}%
}

{
\renewcommand{\arraystretch}{0.7}
\setlength{\tabcolsep}{1.8pt}
\begin{table}[htbp]\footnotesize
  \label{tab:mscoco}
  \centering
  \begin{tabular}{lcccc}
    Metric & Ours(c5) & Base(c5) & Ours(c40) & Base(c40)\\
    \midrule
    BLEU-1 & 0.795 & 0.794 &0.941 &0.939 \\
    BLEU-2 &  0.627& 0.623 &0.869 &0.861\\
    BLEU-3&0.474 & 0.470 &0.764 & 0.754\\
    BLEU-4 & 0.352& 0.349 & 0.647 & 0.637\\
    \midrule
    METEOR & 0.271& 0.268 & 0.355 & 0.351\\
    ROUGE-L & 0.567& 0.564 &0.710 & 0.705\\
    CIDEr-D & 1.100& 1.091 & 1.116 & 1.106\\
    \bottomrule
  \end{tabular}
  \caption{Results on MSCOCO test, reported by the MSCOCO
    server. Ours: ATTN+CIDER+DISC(1). Baseline: ATTN+CIDER.}
  \vspace{-.5em}
\end{table}
}

\section{Improved diversity with discriminability objective}
Table \ref{tab:num_distinct} shows that with larger $\lambda$ we can get longer and more diverse
captions. We also observe pure CIDEr optimization will harm the diversity of output captions.

In Figure \ref{fig:qual}, we show pairs of images that have the same caption
generated by ATTN+CIDER~\cite{rennie2016self}, where our method (ATTN+CIDER+DISC(1)) can generate
more specific captions.

{
\renewcommand{\arraystretch}{0.7}
\begin{table}[htbp]\footnotesize
  \centering
    \begin{tabular}{lcc}
          & \# distinct captions & Avg. length \\
    \midrule
    FC+MLE & 2700  & 8.99 \\
    FC+CIDER\cite{rennie2016self} & 2242  & 9.30 \\
    FC+CIDER+DISC (1) & 3204  & 9.32 \\
    FC+CIDER+DISC (5) & 4379  & 9.45 \\
    FC+CIDER+DISC (10) & 4634  & 9.78 \\
    \midrule
    ATTN+MLE & 2982  & 9.01 \\
    ATTN+CIDER~\cite{rennie2016self} & 2640  & 9.20 \\
    ATTN+CIDER+DISC (1) & 3235  & 9.28 \\
    ATTN+CIDER+DISC (5) & 4089  & 9.54 \\
    ATTN+CIDER+DISC (10) & 4471  & 9.84 \\
    \bottomrule
    \end{tabular}%
  \caption{Distinct caption number on validation set (split in \cite{vedantam2017context}), and average sentence length with different method}
  \label{tab:num_distinct}%
\end{table}%
}

\section{Comparison to \cite{dai2017contrastive,dai2017towards,shetty2017speaking}}
First, while the losses in \cite{dai2017contrastive,dai2017towards,shetty2017speaking} and our
discriminability loss all deal with matched vs. mismatched
image/caption pairs, the actual formulations are very
different.

If $(I,c)$ is the matched image/caption pair, $I',c'$ are
other images/captions, and $\widehat{c}$ the generated caption, the
losses are related to the following notions of ``contrast'' :
\begin{align*}
  &\text{\cite{dai2017contrastive}:}~(I,c)\,\text{vs}\,(I,c'),\\
  &\text{\cite{dai2017towards,shetty2017speaking}}:~(I,c)\,\text{vs}\,(I,\widehat{c}),\,(I,c)\,\text{vs}\,(I,c')\\
  &\text{Ours:}~(I,\widehat{c})\,\text{vs}\,(I,\widehat{c}'),\,
    (I,\widehat{c})\,\text{vs}\,(I',\widehat{c})
\end{align*}

Second, compared to \cite{dai2017towards,shetty2017speaking}, we have a different focus and different evaluation results.
The focus in \cite{dai2017towards,shetty2017speaking} is generating a diverse set of natural captions; in
contrast, we go after discriminability (accuracy) with a single
caption.  We demonstrate improvement in \textbf{both} standard metrics and
our target (discriminability), while they have to sacrifice the former
to improve the performance under their target (diversity).

On SPICE subcategory score, count and size are improved after applying the GAN training in \cite{shetty2017speaking} while our discriminability loss helps color, attribute and count. That implies that our discriminability objective is in favor of different aspects of captions from the discriminator in GAN.

Third, the aims in \cite{dai2017contrastive} are more aligned with ours,
but \cite{dai2017contrastive} does not
explicitly incorporate discriminative task into learning, while we
do. It is also not clear that \cite{dai2017contrastive} could allow for CIDEr optimization,
which we do allow for, and which appears to lead to significant
improvement over MLE.

Finally, none of \cite{dai2017contrastive,dai2017towards,shetty2017speaking} address discriminability of resulting
captions, or report discriminability results comparable to ours.

\begin{figure}[htbp]
\begin{minipage}[t]{.45\linewidth}
\includegraphics[height=3cm]{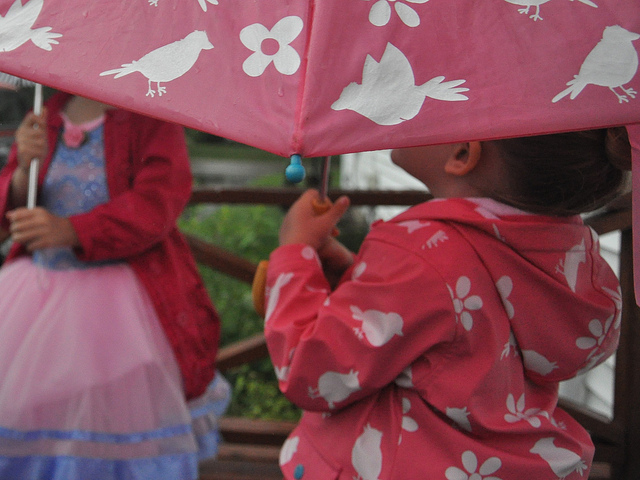}
\setstretch{.75}
{\footnotesize
{\bf Human}: a young child holding an umbrella with birds and flowers
\\
{\bf ATTN+CIDER~\cite{rennie2016self}}: a group of people standing in the rain with an umbrella\\
{\bf Ours}:a little girl holding an umbrella in the rain\\
}
\end{minipage}%
\hspace{1em}
\begin{minipage}[t]{.45\linewidth}
\includegraphics[height=3cm]{images/COCO_train2014_000000231878.jpg}
\setstretch{.75}
{\footnotesize
{\bf Human}: costumed wait staff standing in front of a restaurant awaiting customers\\
{\bf ATTN+CIDER~\cite{rennie2016self}}: a group of people standing in the rain with an umbrella\\
{\bf Ours}: a group of people standing in front of a building\\
}
\end{minipage}
\begin{minipage}[t]{.45\linewidth}
\includegraphics[height=3cm]{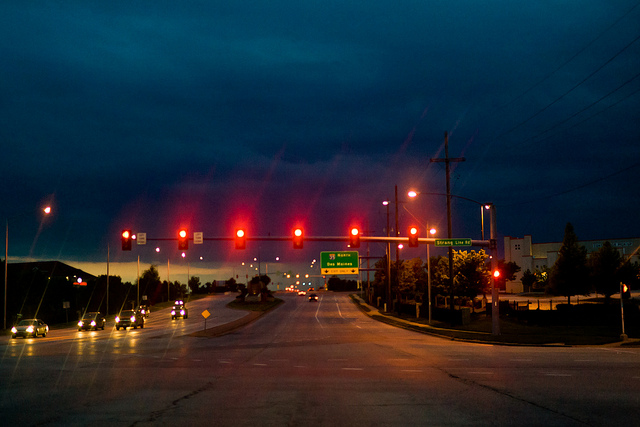}
\setstretch{.75}
{\footnotesize
{\bf Human}: a street that goes on to a high way with the light on red
\\
{\bf ATTN+CIDER~\cite{rennie2016self}}: a traffic light on the side of a city street\\
{\bf Ours}:a street at night with traffic lights at night\\
}
\end{minipage}%
\hspace{1em}
\begin{minipage}[t]{.45\linewidth}
\includegraphics[height=3cm]{images/COCO_train2014_000000554439.jpg}\\
\setstretch{.75}
{\footnotesize
{\bf Human}: view of tourist tower behind a traffic signal\\
{\bf ATTN+CIDER~\cite{rennie2016self}}: a traffic light on the side of a city street\\
{\bf Ours}: a traffic light sitting on the side of a city\\
}
\end{minipage}

\begin{minipage}[t]{.45\linewidth}
\includegraphics[height=3cm]{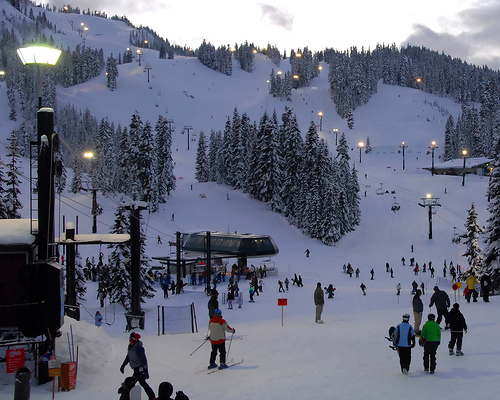}
\setstretch{.75}
{\footnotesize
{\bf Human}: people skiing in the snow on the mountainside
\\
{\bf ATTN+CIDER~\cite{rennie2016self}}: a group of people standing on skis in the snow\\
{\bf Ours}:a group of people skiing down a snow covered slope\\
}
\end{minipage}%
\hspace{1em}
\begin{minipage}[t]{.45\linewidth}
\includegraphics[height=3cm]{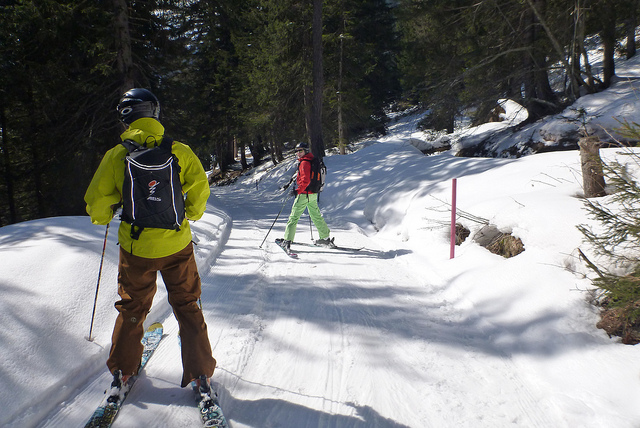}
\setstretch{.75}
{\footnotesize
{\bf Human}: two skiers travel along a snowy path towards trees\\
{\bf ATTN+CIDER~\cite{rennie2016self}}: a group of people standing on skis in the snow\\
{\bf Ours}: two people standing on skis in the snow\\
}
\end{minipage}

\caption{The human caption, caption generated by ATTN+CIDER, and caption generated by 
ATTN+CIDER+DISC(1) (denoted as Ours in the figure) for each image}
\label{fig:qual}
\end{figure}

\section{More results}
Following up the analysis of SPICE sub-scores in the paper, showing improvement due to our proposed objective, 
we show in Figures \ref{fig:color}, \ref{fig:attribute}, \ref{fig:cardinality} some examples demonstrating such improvement on color, attribute and cardinality aspects of the scene. For each figure, the captions below are describing the target images(left) generated from different models (ATTN+MLE and ATTN+CIDER are baselines, and ATTN+CIDER+DISC(x) are our method with different $\lambda$ value). Right images are the corresponding distractors selected in val/test set; these pairs were included in AMT experiments.

Figure \ref{fig:amt} shows some more examples and Figure \ref{fig:failure} failure cases of our methods. We highlight in \textcolor{green!50!black}{green} caption elements that (subjectively) seem to aid discriminability, and in \textcolor{red}{red} the portions that seem incorrect or jarringly non-fluent.

\begin{figure}
\begin{minipage}[t]{\linewidth}
\begin{minipage}[t]{.45\linewidth}
\includegraphics[height=3cm]{images/COCO_val2014_000000358750.jpg}
\end{minipage}%
\hspace{1em}
\begin{minipage}[t]{.45\linewidth}
\includegraphics[height=3cm]{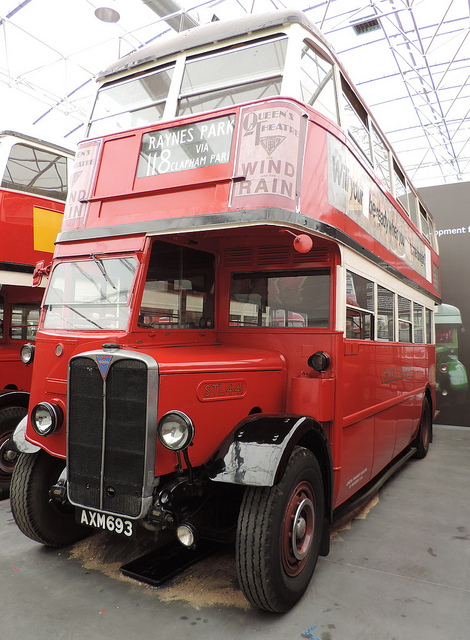}
\end{minipage}
\end{minipage}
\setstretch{.75}
{\footnotesize
{\bf ATTN+MLE}: a double decker bus parked in front of a building
\\
{\bf ATTN+CIDER~\cite{rennie2016self}}: a double decker bus parked in front of a building\\
{\bf ATTN+CIDER+DISC(1)}: a \textcolor{green!50!black}{blue} double decker bus parked in front of a building\\
{\bf ATTN+CIDER+DISC(10)}: a \textcolor{green!50!black}{blue} double decker bus parked in front of a \textcolor{green!50!black}{brick} building\\
}
\begin{minipage}[t]{\linewidth}
\begin{minipage}[t]{.45\linewidth}
\setstretch{.75}
\includegraphics[height=3cm]{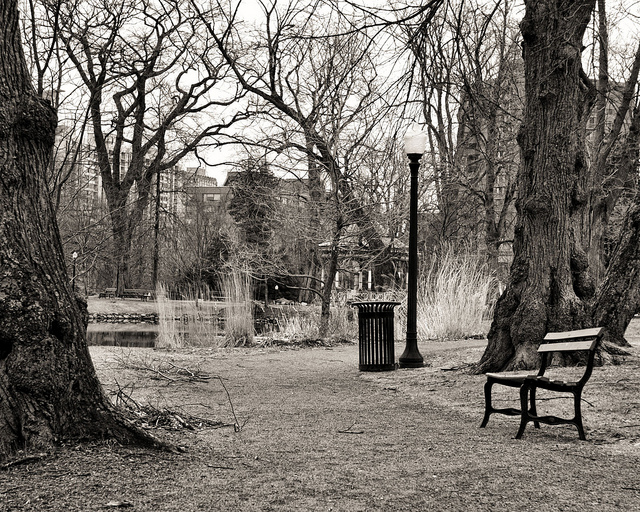}
\end{minipage}%
\hspace{1em}
\begin{minipage}[t]{.45\linewidth}
\includegraphics[height=3cm]{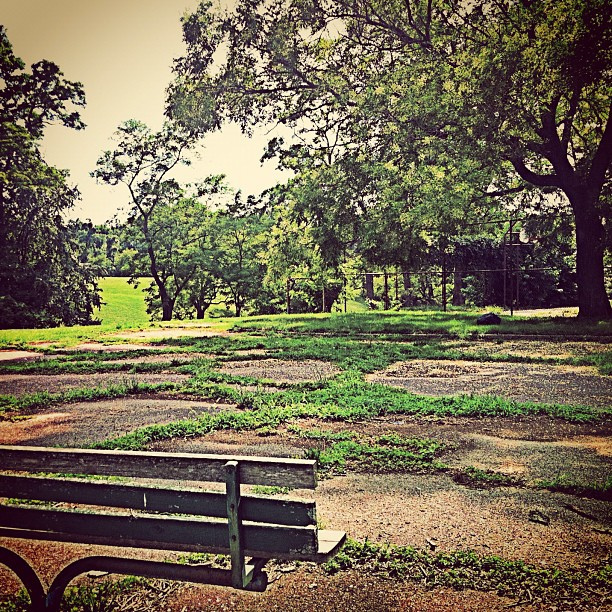}
\end{minipage}
\end{minipage}
{\footnotesize
{\bf ATTN+MLE}: a park bench sitting next to a tree
\\
{\bf ATTN+CIDER~\cite{rennie2016self}}: a bench sitting in the middle of a tree\\
{\bf ATTN+CIDER+DISC(1)}: a \textcolor{green!50!black}{black and white} photo of a bench in the park\\
{\bf ATTN+CIDER+DISC(10)}: a \textcolor{green!50!black}{black and white} photo of a park with a tree\\
}

\begin{minipage}[t]{\linewidth}
\begin{minipage}[t]{.45\linewidth}
\setstretch{.75}
\includegraphics[height=3cm]{images/COCO_train2014_000000047084.jpg}
\end{minipage}%
\hspace{1em}
\begin{minipage}[t]{.45\linewidth}
\includegraphics[height=3cm]{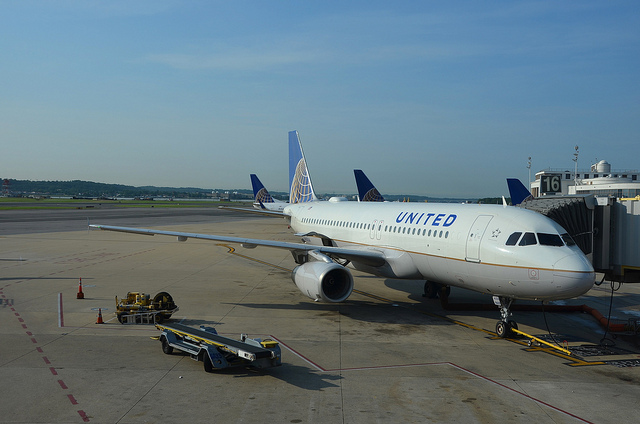}
\end{minipage}
\end{minipage}
{\footnotesize
{\bf ATTN+MLE}: a large jetliner sitting on top of an airport tarmac
\\
{\bf ATTN+CIDER~\cite{rennie2016self}}: a large airplane sitting on the runway at an airport\\
{\bf ATTN+CIDER+DISC(1)}: a \textcolor{green!50!black}{blue} airplane sitting on the tarmac at an airport\\
{\bf ATTN+CIDER+DISC(10)}: a \textcolor{green!50!black}{blue} \textcolor{red}{and blue} airplane sitting on a tarmac with a plane\\
}
\caption{Examples of cases where our method improves the color accuracy of generated captions.}
\label{fig:color}
\end{figure}

\begin{figure}
\begin{minipage}[t]{\linewidth}
\begin{minipage}[t]{.45\linewidth}
\setstretch{.75}
\includegraphics[height=3cm]{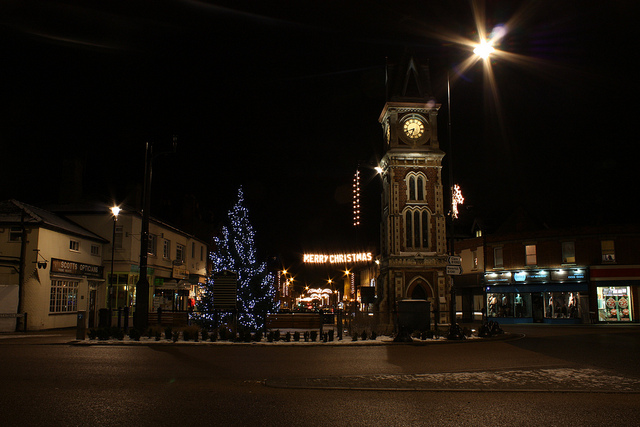}
\end{minipage}%
\hspace{2.5em}
\begin{minipage}[t]{.45\linewidth}
\includegraphics[height=3cm]{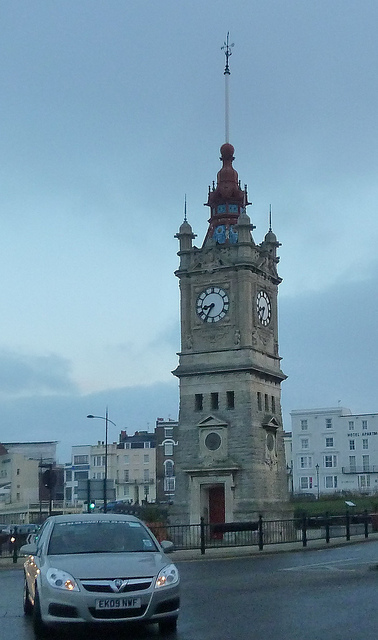}
\end{minipage}
\end{minipage}
\setstretch{.75}
{\footnotesize
{\bf ATTN+MLE}: a large clock tower towering over a city street
\\
{\bf ATTN+CIDER~\cite{rennie2016self}}: a clock tower in the middle of a city street\\
{\bf ATTN+CIDER+DISC(1)}: a city street with a clock tower \textcolor{green!50!black}{at night}\\
{\bf ATTN+CIDER+DISC(10)}: a lit building with a clock tower \textcolor{green!50!black}{at night} \textcolor{red}{at night}\\
}
\begin{minipage}[t]{\linewidth}
\begin{minipage}[t]{.45\linewidth}
\setstretch{.75}
\includegraphics[height=3cm]{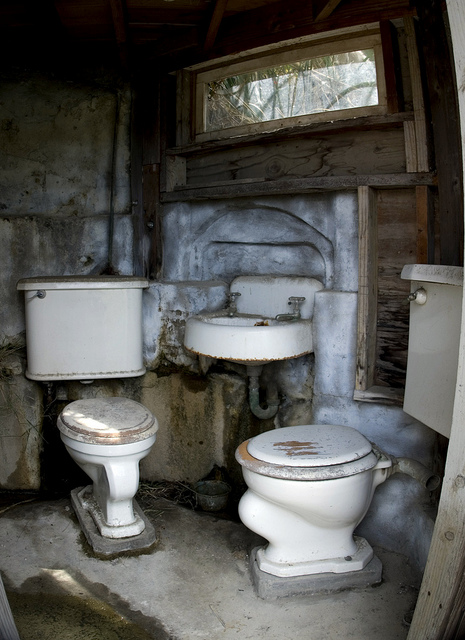}
\end{minipage}%
\hspace{1em}
\begin{minipage}[t]{.45\linewidth}
\includegraphics[height=3cm]{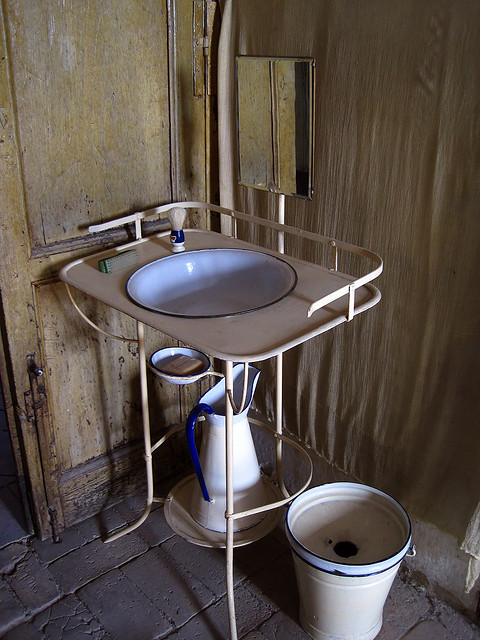}
\end{minipage}
\end{minipage}
\setstretch{.75}
{\footnotesize
{\bf ATTN+MLE}: a bathroom with a toilet and a toilet
\\
{\bf ATTN+CIDER~\cite{rennie2016self}}: a bathroom with a toilet and sink in the\\
{\bf ATTN+CIDER+DISC(1)}: a \textcolor{green!50!black}{dirty} bathroom with a toilet and a window\\
{\bf ATTN+CIDER+DISC(10)}: a \textcolor{green!50!black}{dirty} bathroom with a toilet and \textcolor{red}{a dirty}\\
}
\caption{Examples of cases where our method improves the attribute accuracy of generated captions.}
\label{fig:attribute}
\end{figure}

\begin{figure}
\begin{minipage}[t]{\linewidth}
\begin{minipage}[t]{.45\linewidth}
\setstretch{.75}
\includegraphics[height=3cm]{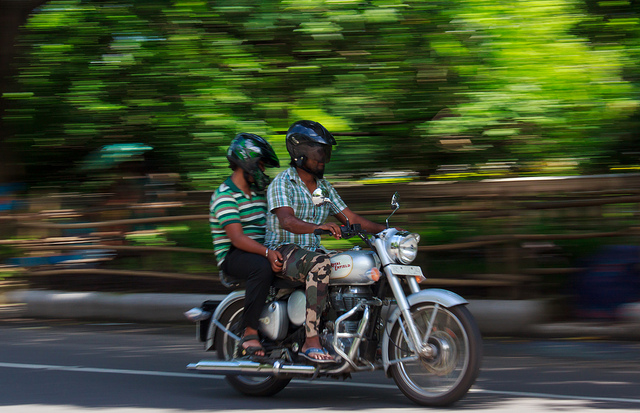}
\end{minipage}%
\hspace{3em}
\begin{minipage}[t]{.45\linewidth}
\includegraphics[height=3cm]{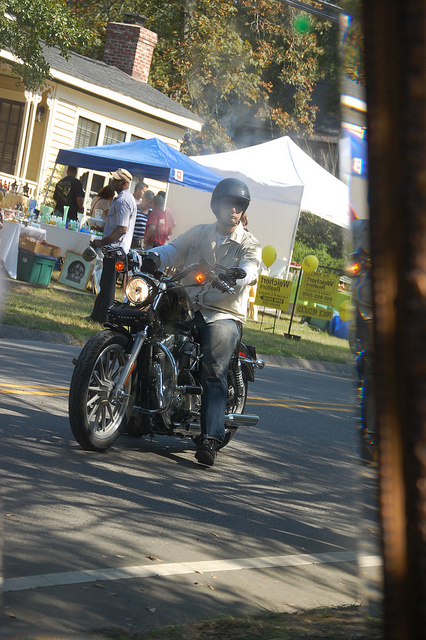}
\end{minipage}
\end{minipage}
\setstretch{.75}
{\footnotesize
{\bf ATTN+MLE}: a couple of people riding on the back of a motorcycle
\\
{\bf ATTN+CIDER~\cite{rennie2016self}}: a man riding a motorcycle on the street\\
{\bf ATTN+CIDER+DISC(1)}: \textcolor{green!50!black}{two people} riding a motorcycle on a city street\\
{\bf ATTN+CIDER+DISC(10)}: \textcolor{green!50!black}{two people} riding a motorcycle on a road\\
}
\begin{minipage}[t]{\linewidth}
\begin{minipage}[t]{.45\linewidth}
\setstretch{.75}
\includegraphics[height=3cm]{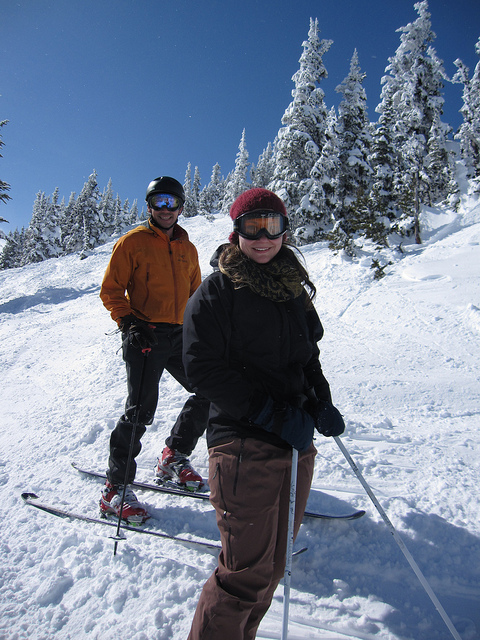}
\end{minipage}%
\hspace{1em}
\begin{minipage}[t]{.45\linewidth}
\includegraphics[height=3cm]{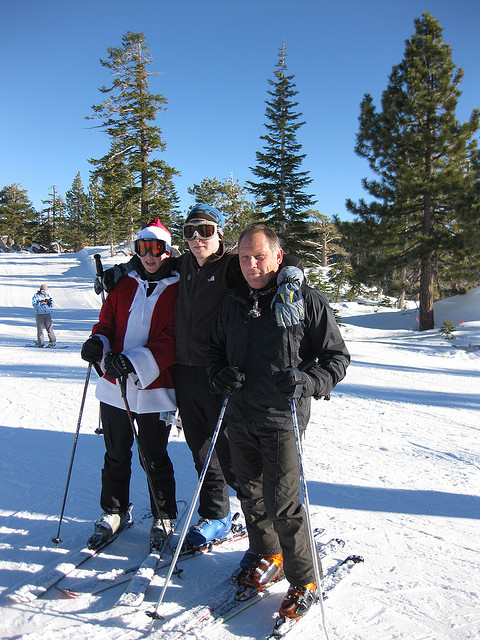}
\end{minipage}
\end{minipage}
\setstretch{.75}
{\footnotesize
{\bf ATTN+MLE}: a couple of people standing on top of a snow covered slope
\\
{\bf ATTN+CIDER~\cite{rennie2016self}}: a couple of people standing on skis in the snow\\
{\bf ATTN+CIDER+DISC(1)}: \textcolor{green!50!black}{two people} standing on skis in the snow\\
{\bf ATTN+CIDER+DISC(10)}: \textcolor{green!50!black}{two people} standing in the snow with a snow\\
}
\caption{Examples of cases where our method improves the cardinality accuracy of generated captions.}
\label{fig:cardinality}
\end{figure}







\begin{figure}
\begin{minipage}[t]{\linewidth}
\begin{minipage}[t]{.45\linewidth}
\includegraphics[height=3cm]{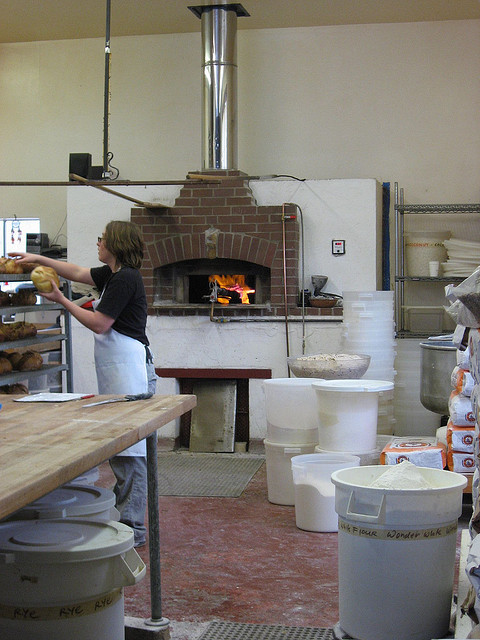}
\end{minipage}%
\hspace{1em}
\begin{minipage}[t]{.45\linewidth}
\includegraphics[height=3cm]{images/COCO_val2014_000000505728.jpg}
\end{minipage}
\end{minipage}
\setstretch{.75}
{\footnotesize
{\bf ATTN+MLE}: a woman standing in a kitchen preparing food
\\
{\bf ATTN+CIDER~\cite{rennie2016self}}: a woman standing in a kitchen preparing food\\
{\bf ATTN+CIDER+DISC(1)}: a woman standing in a kitchen with \textcolor{green!50!black}{a fireplace}\\
{\bf ATTN+CIDER+DISC(10)}: a woman standing in a kitchen with \textcolor{green!50!black}{a brick oven}\\
}

\begin{minipage}[t]{\linewidth}
\begin{minipage}[t]{.45\linewidth}
\includegraphics[height=3cm]{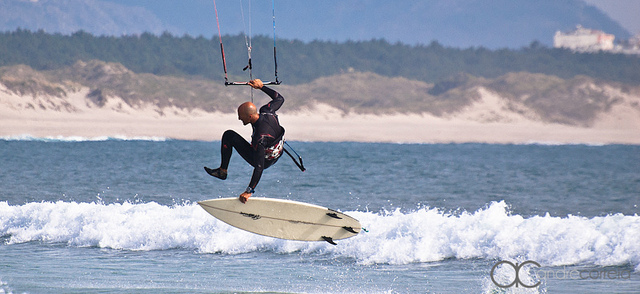}
\end{minipage}%
\hspace{1em}
\begin{minipage}[t]{.45\linewidth}
\includegraphics[height=3cm]{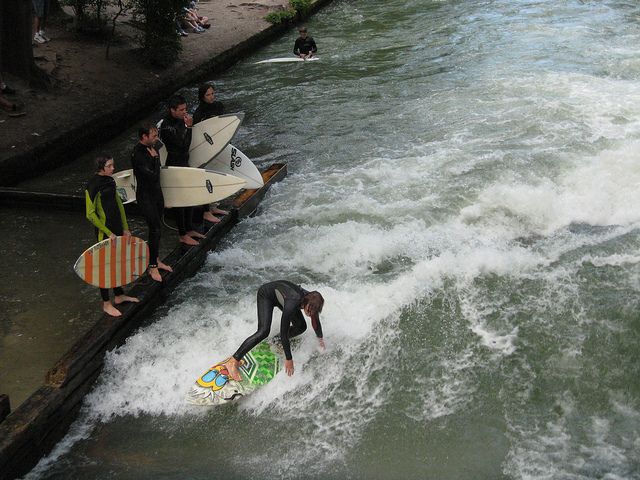}
\end{minipage}
\end{minipage}
\setstretch{.75}
{\footnotesize
{\bf ATTN+MLE}: a man on a surfboard in the water
\\
{\bf ATTN+CIDER~\cite{rennie2016self}}: a man riding a wave on a surfboard in the ocean\\
{\bf ATTN+CIDER+DISC(1)}: a man riding a \textcolor{green!50!black}{kiteboard} \textcolor{red}{on the ocean in the ocean}\\
{\bf ATTN+CIDER+DISC(10)}: a man \textcolor{green!50!black}{kiteboarding} \textcolor{red}{in the ocean on a ocean}\\
}

\begin{minipage}[t]{\linewidth}
\begin{minipage}[t]{.45\linewidth}
\includegraphics[height=3cm]{images/COCO_train2014_000000493692.jpg}
\end{minipage}%
\hspace{1em}
\begin{minipage}[t]{.45\linewidth}
\includegraphics[height=3cm]{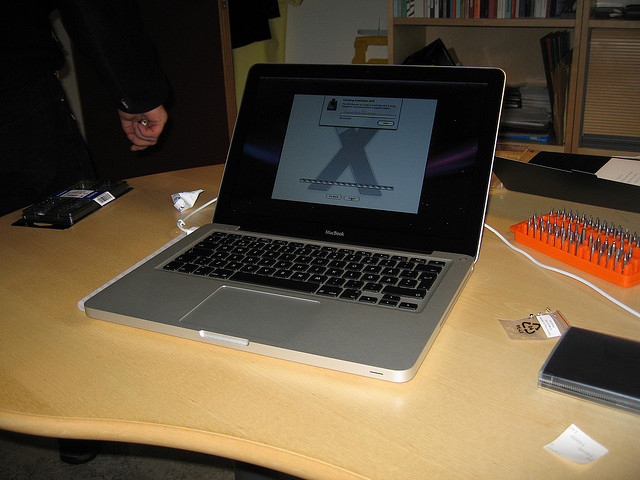}
\end{minipage}
\end{minipage}
\setstretch{.75}
{\footnotesize
{\bf ATTN+MLE}: a room with a laptop and a laptop
\\
{\bf ATTN+CIDER~\cite{rennie2016self}}: a laptop computer sitting on top of a table\\
{\bf ATTN+CIDER+DISC(1)}: a room with a laptop computer and \textcolor{green!50!black}{chairs} in it\\
{\bf ATTN+CIDER+DISC(10)}: a room with a laptops and \textcolor{green!50!black}{chairs} in a building\\
}

\begin{minipage}[t]{\linewidth}
\begin{minipage}[t]{.45\linewidth}
\includegraphics[height=3cm]{images/COCO_train2014_000000388031.jpg}
\end{minipage}%
\hspace{1em}
\begin{minipage}[t]{.45\linewidth}
\includegraphics[height=3cm]{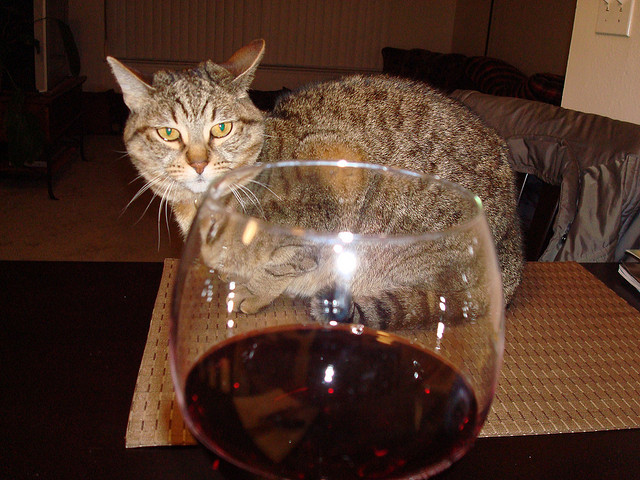}
\end{minipage}
\end{minipage}
\setstretch{.75}
{\footnotesize
{\bf ATTN+MLE}: a cat sitting next to a glass of wine
\\
{\bf ATTN+CIDER~\cite{rennie2016self}}: a cat sitting next to a glass of wine\\
{\bf ATTN+CIDER+DISC(1)}: a cat sitting next to a \textcolor{green!50!black}{bottles} of wine\\
{\bf ATTN+CIDER+DISC(10)}: a cat sitting next to a \textcolor{green!50!black}{bottles} of wine \textcolor{red}{bottles}\\
}
\caption{Some more examples}
\label{fig:amt}
\end{figure}

\begin{figure}

\begin{minipage}[t]{\linewidth}
\begin{minipage}[t]{.45\linewidth}
\setstretch{.75}
\includegraphics[height=3cm]{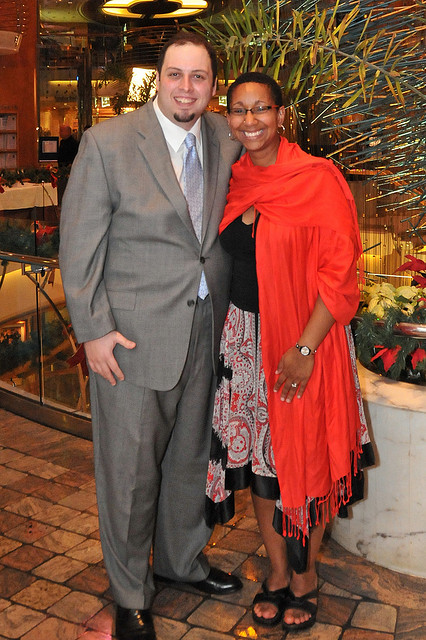}
\end{minipage}%
\hspace{1em}
\begin{minipage}[t]{.45\linewidth}
\includegraphics[height=3cm]{images/COCO_train2014_000000428050.jpg}
\end{minipage}
\end{minipage}
\setstretch{.75}
{\footnotesize
{\bf ATTN+MLE}: a couple of people standing next to each other
\\
{\bf ATTN+CIDER~\cite{rennie2016self}}: a man and a woman standing in a room\\
{\bf ATTN+CIDER+DISC(1)}: a man and a woman standing in a room\\
{\bf ATTN+CIDER+DISC(10)}: \textcolor{red}{two men} standing in a room with a \textcolor{red}{red tie}\\
}
\begin{minipage}[t]{\linewidth}
\begin{minipage}[t]{.45\linewidth}
\setstretch{.75}
\includegraphics[height=3cm]{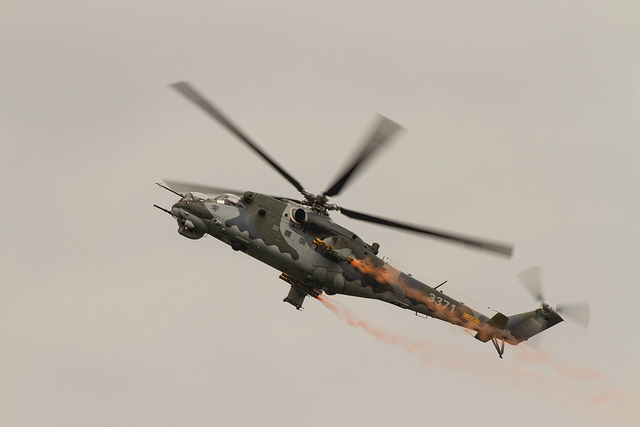}
\end{minipage}%
\hspace{1em}
\begin{minipage}[t]{.45\linewidth}
\includegraphics[height=3cm]{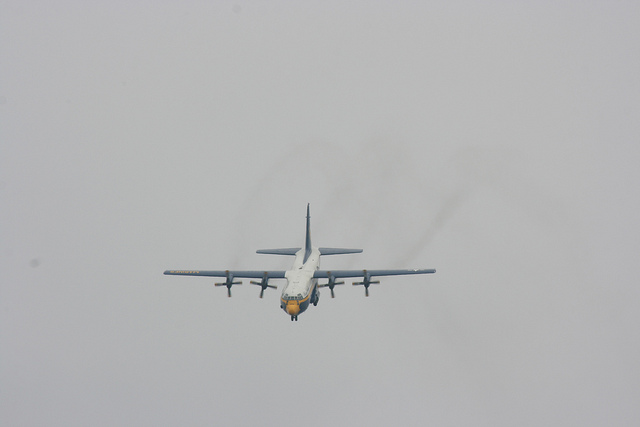}
\end{minipage}
\end{minipage}
\setstretch{.75}
{\footnotesize
{\bf ATTN+MLE}: a helicopter that is flying in the sky
\\
{\bf ATTN+CIDER~\cite{rennie2016self}}: a helicopter is flying in the sky\\
{\bf ATTN+CIDER+DISC(1)}: \textcolor{red}{a fighter jet} flying in the sky\\
{\bf ATTN+CIDER+DISC(10)}: \textcolor{red}{a fighter plane} flying in the sky with smoke\\
}
\begin{minipage}[t]{\linewidth}
\begin{minipage}[t]{.45\linewidth}
\setstretch{.75}
\includegraphics[height=3cm]{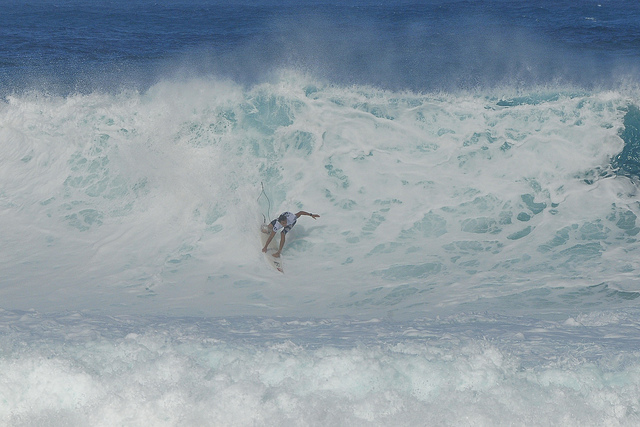}
\end{minipage}%
\hspace{1em}
\begin{minipage}[t]{.45\linewidth}
\includegraphics[height=3cm]{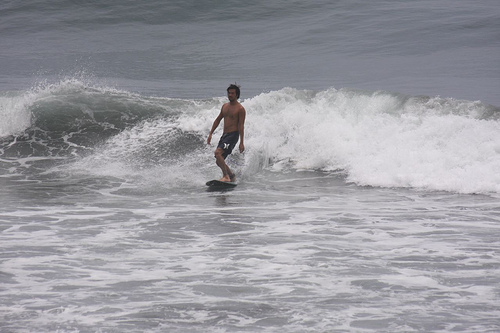}
\end{minipage}
\end{minipage}
\setstretch{.75}
{\footnotesize
{\bf ATTN+MLE}: a man riding a wave on top of a surfboard
\\
{\bf ATTN+CIDER~\cite{rennie2016self}}: a man riding a wave on a surfboard in the ocean\\
{\bf ATTN+CIDER+DISC(1)}: a person riding a wave on a surfboard in the ocean\\
{\bf ATTN+CIDER+DISC(10)}: a person  \textcolor{red}{kiteboarding} on a wave in the ocean\\
}
\caption{Some failure cases of our algorithm}
\label{fig:failure}
\end{figure}
\fi
\end{document}